\documentclass{article} 
\usepackage{iclr2022_workshop,times}
\usepackage{multirow}

\usepackage{amsmath,amsfonts,bm}

\def\eqref#1{equation~\ref{#1}}

\def\1{\bm{1}}

\DeclareMathAlphabet{\mathsfit}{\encodingdefault}{\sfdefault}{m}{sl}
\SetMathAlphabet{\mathsfit}{bold}{\encodingdefault}{\sfdefault}{bx}{n}

\usepackage{hyperref}
\usepackage{url}
\usepackage{microtype}
\usepackage{graphicx}
\usepackage{subfigure}
\usepackage{algorithm}
\usepackage[algo2e,ruled,linesnumbered,lined]{algorithm2e}
\usepackage{booktabs} 
\graphicspath{ {./images/} }
\usepackage{hyperref}

\usepackage{amsmath}
\usepackage{amssymb}
\usepackage{mathtools}
\usepackage{amsthm}

\usepackage[capitalize,noabbrev]{cleveref}

\theoremstyle{plain}

\theoremstyle{definition}

\theoremstyle{remark}

\usepackage[textsize=tiny]{todonotes}
\newcommand{\simm}{\texttt{sim}}
       
\title{Local Learning with Neuron Groups}
\author{\parbox{\textwidth}{\centering
    Adeetya Patel$^{1,2}\;$\hspace{-10pt}
    \qquad Michael Eickenberg$^{3 \; }$ \hspace{-10pt}
    \qquad Eugene Belilovsky$^{1,2}$} \vspace{5pt}\\
\centerline{$^1$~Concordia University\hspace{1em} $^2$~Mila -- Quebec AI Institute\hspace{1em} $^3$~Flatiron Institute}\\
\centerline{{\tt\small \{adeetya.patel, eugene.belilovsky\}@concordia.ca, }} \\
\centerline{{\tt\small 
meickenberg@flatironinstitute.org}}
}

\iclrfinalcopy 
\begin{document}

\maketitle

\begin{abstract}
Traditional deep network training methods optimize a monolithic objective function jointly for all the components. This can lead to various inefficiencies in terms of potential parallelization. Local learning is an approach to model-parallelism that removes the standard end-to-end learning setup and utilizes local objective functions to permit parallel learning amongst model components in a deep network. Recent works have demonstrated that variants of local learning can lead to efficient training of modern deep networks. However, in terms of how much computation can be distributed, these approaches are typically limited by the number of layers in a network. In this work we propose to study how local learning can be applied at the level of splitting layers or modules into sub-components, adding a notion of width-wise modularity to the existing depth-wise modularity associated with local learning. We investigate local-learning penalties that permit such models to be trained efficiently. Our experiments on the CIFAR-10, CIFAR-100, and Imagenet32 datasets demonstrate that introducing width-level modularity can lead to computational advantages over existing methods based on local learning and opens new opportunities for improved model-parallel distributed training. Code is available at: \url{https://github.com/adeetyapatel12/GN-DGL}.
\end{abstract}

\section{Introduction}
Neural networks are typically trained by stochastic gradient descent in combination with the back-propagation algorithm \citep{krizhevsky2012imagenet, huang2019convolutional, simonyan2014very, szegedy2015going}. This learning algorithm allows joint adaptation of all layers and neuron connections in a network based on a global objective function. It is typically believed that this joint adaptation is essential to obtain a high performance for large-scale data sets as joint adaptation towards an overarching objective allows the individual layers and neurons to adapt their functionality efficiently. 
On the other hand, several recent works have studied the idea to use a purely local loss function, where there is no feedback between the independent layers.
These approaches are based on the classic sequential greedy learning procedure \citep{ivakhnenko,bengio2007greedy} but allow the different model components to simultaneously learn their model parameters. Surprisingly, relying purely on this forward communication, high-performance models can be constructed \citep{pmlr-v119-belilovsky20a, nokland2019training}. These observations have implications on both the functional properties of high-performance deep networks as well as practical implications for distributed training of neural networks. If layerwise learning can still bring high performance, this raises the question: to what degree do neural network components need to be trained jointly?


We can obtain some inspiration from biological neural systems. Biological neural systems rely on extremely localized synaptic updates, often modeled as Hebbian learning \citep{hebb1949organization} or adaptations thereof. The primary ways for a biological neural system to incorporate global information into learning is by way of feedback/recurrent neural connections, which can bring information that was computed at a later stage back to earlier neurons and hence be incorporated by local learning. 
Local learning in biological systems includes spatially local processing in retinotopic maps, which inspired locally connected architectures such as convolutional networks. Further, in biological systems, there often exist parallel pathways that perform different but somewhat overlapping computations (e.g. magno- and parvocellular visual pathways). We take inspiration from this idea and translate it to a channelwise splitting of computations such that the different components can be efficiently hosted on different processing units.

In this work we 
investigate the limits of using supervised local loss functions, moving them from the layer level to the ``neuron group" level. Specifically, we consider the case where non-overlapping groups of neurons within a network layer each have their own local objective function and optimize their parameters in isolation. Our investigation reveals that networks can still demonstrate high performance despite this modification. We also investigate how to encourage these decoupled networks to learn diverse behavior (i.e. learn different representations from each other) based solely on forward communication. We illustrate the practical applications of this by comparing the proposed method to existing local learning approaches in the context of total training time, inference time, and model performance.


\begin{figure*}[t]
    \centering
    \includegraphics[width=0.65\textwidth]{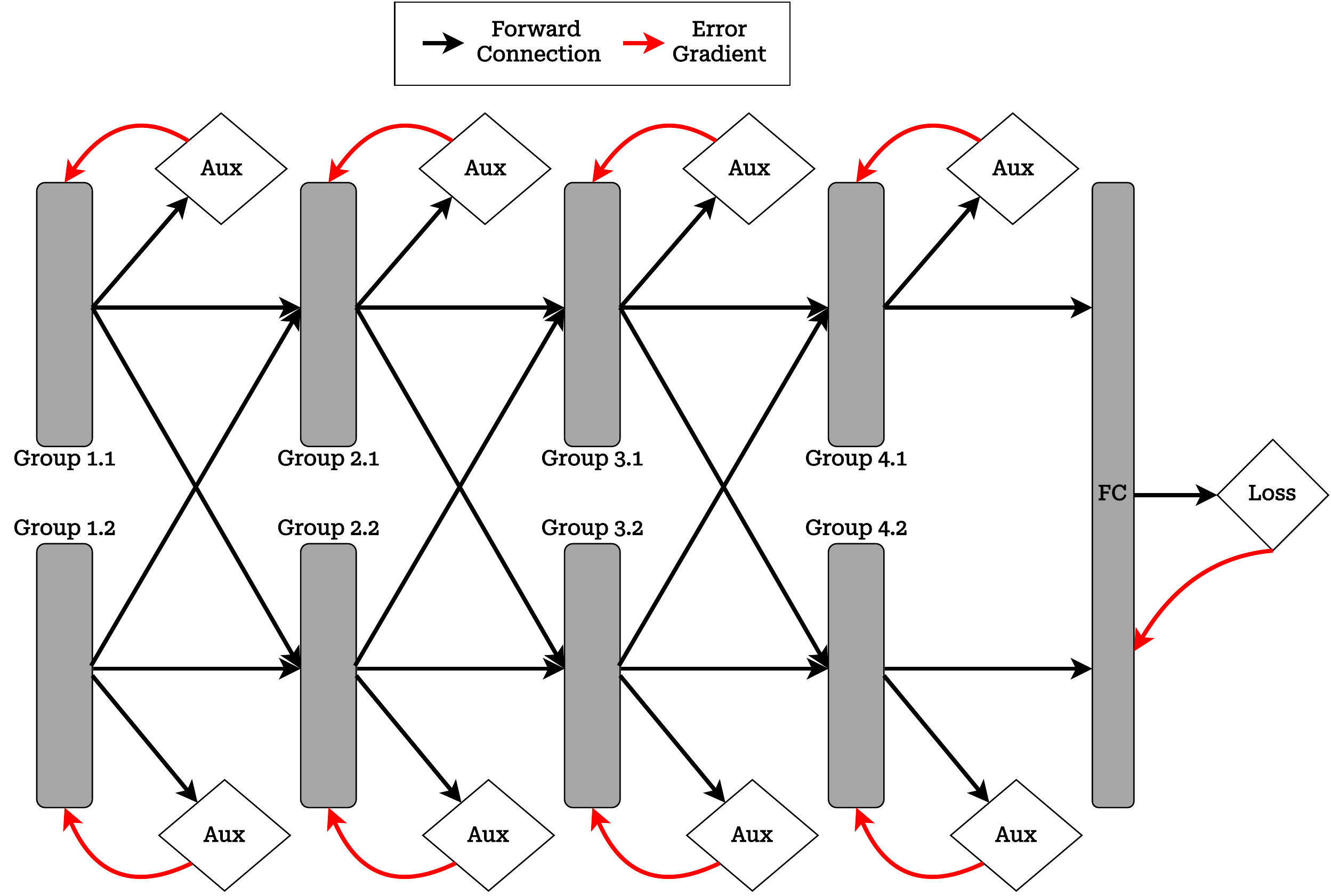}
    \caption{Grouped Neuron DGL, each layer and neuron group has a local objective. All local groups are learned in parallel, feeding their output directly to the next layer's groups. } \label{fig:GN_DGL} 
\end{figure*}
\section{Related Work}
Sequential local learning, where a network is built up through greedily adding individual layers and solving local layerwise optimization problems, has been studied in a number of classical works \citep{ivakhnenko,fahlman1989cascade}. These kinds of approaches were commonly used on simple datasets particularly in the case of unsupervised models \citep{bengio2007greedy,vincent2010stacked}. Their use was motivated primarily by difficulties associated with joint optimization of deep networks such as the vanishing gradient problem. However, such methods fell out of favor with the advent of modern techniques to tackle joint deep network training such as improved initialization, residual connections \citep{he2016deep} and normalization \citep{ioffe2015batch}. These sequential local learning techniques have been revisited recently by \cite{shallow} which showed that this approach can yield high-performance models on large datasets and architectures. \cite{pmlr-v119-belilovsky20a,nokland2019training} further demonstrated that this can be performed in the parallel local learning setting, where layers are learned simultaneously, allowing for improved distributed training of deep networks. It can even be done in an asynchronous manner if each layer is allowed to maintain a memory. 
\cite{wang2021revisiting} further extended this idea, illustrating a novel regularization term that combats the collapsing of the representation towards the target supervised task.  
\cite{laskin2021parallel} has performed a study comparing a number of local learning techniques including interlocking backpropagation on several common large scale tasks.  

Related to our work \cite{veness2019gated} studied a local objective where each neuron solves a binary classification problem. Results were illustrated in an online-learning setting but have not been extended to the standard offline learning settings or to complex datasets such as imagenet. 
\cite{choromanska2019beyond, lee2015difference, pmlr-v119-belilovsky20a} considers local objective functions with targets generated as a function of the global loss. These, however, require feedback communication between the various layers and model components. Our work on the other hand focuses on local learning objectives and their extensions.
\cite{jaderberg2016decoupled} proposed to also use auxiliary networks to locally approximate a gradient in order to allow for parallel learning of neural network layers and blocks. 

Pipelining \citep{huang2018gpipe} is a systems-level solution that can alleviate issues introduced by backpropagation that prevent parallel training at different modules. However, these techniques do not remove the fundamental limits (often described as locks in \cite{jaderberg2016decoupled}) and thus does not allow for a full parallelization.


\begin{figure*}[t]
    \centering
    \includegraphics[width=0.83\textwidth]{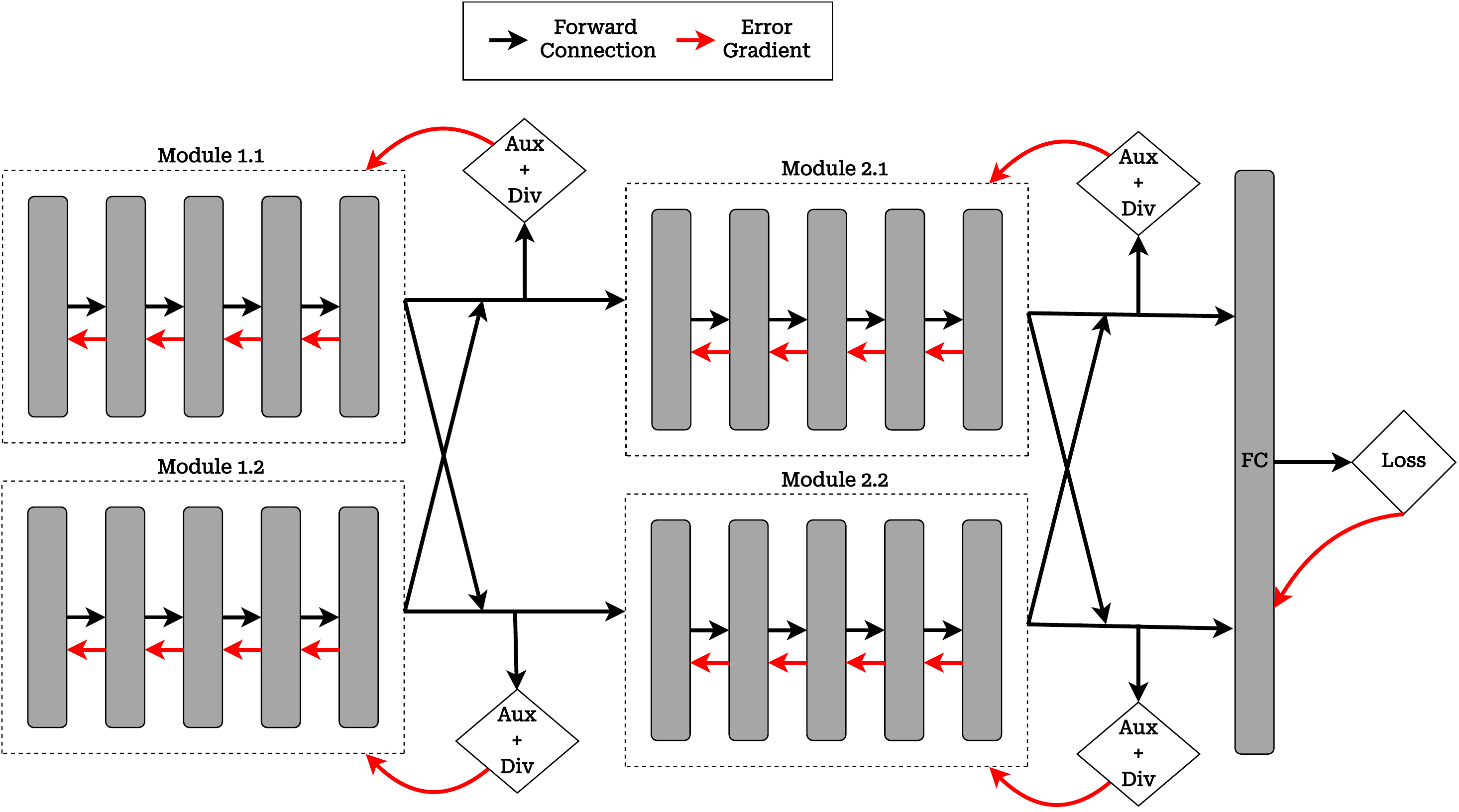}
    \caption{Illustration of our approach. Multi-layer modules are used in combination with a gradient decoupled output and diversity-promoting penalty.} \label{fig:ML_GN_DGL} 
\end{figure*}

\section{Methods}

\subsection{Background on DGL}
We now describe the local learning framework of \cite{pmlr-v119-belilovsky20a,nokland2019training} and introduce notation. Consider an input to a neural network $x_0$, we denote the operations of layer $j$ as $f_{\theta_j}(x_{j-1})$, where $\theta_j$ corresponds to the parameters of the network and $\mathcal{L}(y,x_j;\gamma_j,\theta_j)$ corresponds to local loss function applied to the representation $x_j$, where $\gamma_j$ are parameters of an auxiliary network. In \cite{pmlr-v119-belilovsky20a,nokland2019training}, it is proposed to learn the parameters $\theta_j$ jointly and in parallel. In the next section, we consider further dividing the objective to introduce a group-wise local loss  $\mathcal{L}(y,x_j^i;\gamma_j^i,\theta_j^i)$, where $x_j^i$ is a subset of $x_j$.

\subsection{Grouped Neuron DGL (GN-DGL)}
Building on layerwise local learning we consider further splitting each layer into groups with isolated losses. 
We propose to learn each of these modules in parallel online (as in \cite{pmlr-v119-belilovsky20a, nokland2019training} instead of sequentially as in \cite{shallow}). Specifically, we try to optimize a greedy objective for each of the neuron groups as shown in Algorithm \ref{algo:basic}. Here, each neuron group has approximately the same number of neurons. It is important to note here that during a forward pass each neuron group $i$ at layer $j$ receives as input the output representations, $x_{j-1}^k$ from all groups of the previous layer $j-1$ (i.e. all $k$ in layer $j-1$). 
This is illustrated in Figure \ref{fig:GN_DGL}. We refer to this approach as Grouped Neuron DGL (GN-DGL). Note that in our work the last fully connected layer is also decoupled from previous layer neuron groups.

In \cite{belilovsky2019greedy} the performance of the layerwise models is enhanced by weighted ensembling layerwise predictions. Similarly we can apply an ensembling over both layers and submodules, to obtain final predictions. 


\subsection{Stop-Gradient Grouped Neuron DGL}
\label{stop-grad}
We can loosen the communication 
restrictions
between layer groups, permitting them to send their outputs to each other's auxiliary networks. In this method, depicted in Algorithm~\ref{algo:basic_2}, the auxiliary networks of each neuron group collect the gradient-decoupled output representations from other neuron groups of the same layer and uses it as part of their final prediction. We note that the communication cost can be decreased by spatially pooling the output before sending it to auxiliary networks.
Importantly, this is only a forward connection: no information is sent back to the origin nodes.

\setlength{\textfloatsep}{1pt}
\vspace{10pt}
\begin{algorithm2e}[H]\small
    \caption{Grouped-Neuron DGL}\label{algo:basic}
  \SetAlgoLined
  \DontPrintSemicolon
\KwIn{Mini-batches $\mathcal{S}\triangleq\{(x_0^t,y^t)\}_{t\leq T}$}
\textbf{Initialize} Parameters $\{\theta^i_j,\gamma^i_j\}_{j\leq J, i\leq G}$.\;
\For { $(x_0,y) \in \mathcal{S}$ }
{
\For {$j \in 1,..., J$}
   {
  \For {$i \in 1,..., G$}
   {
   $x^i_j \leftarrow f_{\theta_{j}^i}(x_{j-1})$.\;
   Compute $\nabla_{(\gamma_j^i,\theta_j^i)}\hat{ \mathcal{L}}(y,x^i_j;\gamma_j^i,\theta_j^i)$.\;
$(\theta_j^i,\gamma_j^i)\leftarrow$Update params $(\theta_j^i,\gamma_j^i)$.
}
$x_j \leftarrow \big\Vert_{i=1}^G x^i_j$.\;
   }
 }
\end{algorithm2e}
\vspace{5pt}
\setlength{\textfloatsep}{1pt}
\begin{algorithm2e}[H]\small
    \caption{Stop-Gradient GN-DGL}\label{algo:basic_2}
  \SetAlgoLined
  \DontPrintSemicolon
\KwIn{Mini-batches $\mathcal{S}\triangleq\{(x_0^t,y^t)\}_{t\leq T}$}
\textbf{Initialize} Parameters $\{\theta^i_j,\gamma^i_j\}_{j\leq J, i\leq G}$.\;
\For { $(x_0,y) \in \mathcal{S}$ }
{
\For {$j \in 1,..., J$}
   {
  \For {$i \in 1,..., G$}
   {
   $x^i_j \leftarrow f_{\theta_{j}^i}(x_{j-1})$.\;
   $z^i_j=sg\big[\big\Vert_{k \neq i} x^k_j\big]$ \;
   Compute$\nabla_{(\gamma_j^i,\theta_j^i)}\hat{\mathcal{L}}(y,x^i_j,z^i_j;\gamma_j^i,\theta_j^i)$.\;
$(\theta_j^i,\gamma_j^i)\leftarrow$Update params $(\theta_j^i,\gamma_j^i)$.
}
$x_j \leftarrow \big\Vert_{i=1}^G x^i_j$.\;
   }
 }
\end{algorithm2e}

\subsection{Grouped Neuron DGL with diversity-promoting penalty}
\label{div}
An expected weakness of local group losses is that representations learned in different groups will 
be correlated and redundant to each other, not permitting optimal aggregation of predictive power.
We attempt to address this by introducing a diversity-promoting penalty that produces negligible overhead in communication between modules. Specifically, we allow auxiliary modules to send their softmax output distributions to each other. Our diversity-promoting term in the loss function encourages diversity locally, based solely on communication of softmax layer outputs between auxiliary modules.

For the diversity-promoting penalty, we utilize a variant of the penalty studied in \cite{dvornik2019diversity}. Each neuron group of the network and their corresponding auxiliary modules are parameterized by $\theta^i_j$ and $\gamma^i_j$ respectively, where $i$ and $j$  indicates the group index and layer index respectively. Each group leads to the class probabilities $p^i_j=\textrm{softmax} {(f_{(\theta^i_j, \gamma^i_j)}(x_{j-1})})$. Now consider $\hat{p}^i_j= \frac{p^i_j\odot m_y}{\|p^i_j\odot m_y\|_1}$ where $m_y$ is a vector with zero at position $y$ and 1 otherwise. 
Now, for a group $a$ in the layer $j$, we compute the diversity-promoting penalty with each remaining group $b$ in the same layer as shown in Eq. \ref{eq:div_penalty},

\begin{equation}\label{eq:div_penalty}
 \phi(\hat{p}^a_j, \hat{p}^b_j) = \simm(\hat{p}^a_j, sg[\hat{p}^b_j])
\end{equation}

Here, $sg$ denotes the stop-gradient operation and $\simm$ is the cosine similarity. It is important to note here that to estimate the diversity-promoting penalty for group $a$ with all remaining groups $b$ of the same layer, we use a gradient-decoupled version of $\hat{p}^b_j$. This means that no gradient signal needs to be sent back. This penalty encourages diversity by pushing the non-target-label softmax activations of the different groups away from each other.

\paragraph{Layerwise ensemble evaluation}\label{layerwise-ens-eval} Typically, the prediction of a neural network is given by the softmax outputs of the last layer of the network. Since in this local learning paradigm we train each group in isolation with separate local loss functions, we can also aggregate a number of predictors at different depths as well as for different neuron groups. Following \cite{belilovsky2019greedy} we investigate how each group performs on the task and evaluate ensembling these neuron groups to measure the performance of the network.

Consider the vector of outputs for the softmax layer $\bm{l}_{ij}$ produced by each neuron group, where $i$ and $j$ represents the index of a neuron group over width and depth respectively. To obtain the final output we take a weighted mean of all vectors by assigning increasing weights over depth and keeping similar weights for the groups at same depth. 
\[l_e = \displaystyle
\dfrac{1}{K \times G} \sum_{i=1}^{G} \sum_{j=1}^{K}\alpha_{ij}\bm{l}_{ij}\]
We keep the $\alpha_{ij}$ constant
for the groups of a given layer. We vary them  for different layers $j$ using the weighting given in \cite{belilovsky2019greedy}. 

\section{Experiments and Results}
Our experimental results start with an initial set of ablations and then focus on evaluations of performance tradeoffs in model-parallel learning setting.
First, we study a simple VGG6a model using layerwise and group-neuron level local loss functions. This allows us to establish basic performance characteristics of this approach compared to layerwise training.
In the second set of experiments, we focus on an application to model-parallel training, comparing our method in the case of multi-layer modules to the recent methods such as Decoupled Greedy Learning (DGL) \cite{pmlr-v119-belilovsky20a} and local learning with Information Propagation (InfoPro) \cite{wang2021revisiting}. 


\paragraph{Datasets} 
We perform ablations using the popular CIFAR-10 dataset used in many prior works \citep{pmlr-v119-belilovsky20a,Huo2018} in order to obtain intuition on the performance of the various components. Our experiments studying trade-offs in model-parallelism criteria utilize CIFAR-10, CIFAR-100, and a downsampled version of the Imagenet dataset (denoted Imagenet32 \cite{chrabaszcz2017downsampled}).

\paragraph{Training hyper-parameters}
For experiments in Sec \ref{sec:gn-dgl}, the networks are trained using the Adam optimizer \citep{kingma2014adam} and a batch size of 128 for total 150 epochs. We use the initial learning rate of 0.001 with a decay at epochs 50 and 100 to 0.0005 and 0.0001 respectively with a dropout rate of 0.01

For the experiments in Sec \ref{sec:ml-gn-dgl} we use the hyper-parameters from \cite{wang2021revisiting}. For all datasets, we train the networks using an SGD optimizer with a Nesterov momentum of 0.9 for 160 epochs. We use an initial learning rate of 0.8 with cosine learning rate annealing. The batch size is set to 1024 and $\ell_2$ weight decay ratio of 1e-4 is adopted. 

\begin{table*}[!h]
\centering
\begin{tabular}{|c|c|c|c|c|c|c|}
\hline
\textbf{Method} & DGL & 2-Group & 4-Group & 6-Group & 8-Group & 10-Group \\
\hline
\textbf{Test Accuracy} & 92.25 & 91.63 & 90.55 & 89.8 & 89.07 & 88.91 \\
\hline
\end{tabular}
 \caption{Grouping of neurons in each layer of VGG6a. We observe with increased number of groups, there is performance degradation, however a surprisingly high overall accuracy can be maintained despite a lack of communication across neuron groups. We note that the presented results do not utilize stop-gradient techniques, nor a local diversity-promoting penalty.} \label{table:vgg6a_naive_splitting} 
\end{table*}

\begin{table*}[!h]
\centering
\begin{tabular}{|c|c|c|c|c|}
\hline
\multirow{3}{*}{\textbf{Method}} & \multirow{3}{*}{DGL} &
\multirow{3}{*}{6-Group GN-DGL} & 6-Group & 6-Group \\
& & & GN-DGL & GN-DGL \\
& &  & with diversity & Stop Gradient + diversity \\
\hline
 \textbf{Test Accuracy} & 92.2  & 89.8 & 90.6 & 91.9 \\
\hline

\end{tabular}
\caption{Ablating the effect of local diversity-promoting penalty and use of stop-gradient communication in GN-DGL.\vspace{7pt}} \label{table:vgg6a_ablation-1} 
\end{table*}
\subsection{Layerwise and Grouped Neuron DGL}
\label{sec:gn-dgl}
The VGGNet used in the experiments, denoted VGG6a, consists of six layers (four convolutional and two fully connected) and is taken from \cite{nokland2019training}. The convolutional layers have 128, 256, 512, 512 channels respectively, and 8192, 1024 features in the last two fully connected layers. To report DGL results, we train each layer with its own auxiliary network as described in \cite{pmlr-v119-belilovsky20a}. For GN-DGL experiments, we further split the layers widthwise into 2 to 10 neuron groups. The neuron groups within each layer and their auxiliary networks are trained locally by the auxiliary loss. The details about widthwise splitting are further elaborated below at the end of Sec \ref{splitting_config_vgg}. Each neuron group has its own auxiliary loss except the last fully connected layer, as shown in Figure \ref{fig:GN_DGL}. We use the same auxiliary network design and loss (termed \textit{predsim}) as \cite{nokland2019training}.

We first consider naive splitting (GN-DGL without stop-gradient and diversity). Results of naive splitting experiments on VGG6a is presented in Table \ref{table:vgg6a_naive_splitting}. Here the increasing number of groups are indicated by G, the minimum number of groups in any layer. We also report the DGL \citep{pmlr-v119-belilovsky20a} test accuracy for VGG6a, where the network is only divided depthwise and each layer is a local module in itself with their own auxiliary network and loss. We observe that as we increase $G$, the test accuracy slightly decreases (which is expected). We believe the decrease in accuracy is mainly due to correlated feature, arising from each neuron group being trained in isolation.

\begin{table}[!h]
\centering
\begin{tabular}{|c|c|c|c|} 
\hline
& \textbf{VGG6a} & \textbf{2xVGG6a} & \textbf{3xVGG6a}\\ 
\hline
DGL & 92.25 &  93.27 & 93.27 \\ 
\hline
6-Group GN-DGL & 89.8 & 92.6 & 93.28 \\ 

\hline
\end{tabular}
\caption{Ablating the effect of width in GN-DGL. We observe that as width increases, using GN-DGL begins to yield performance comparable to DGL, with increased parallelization.\vspace{-5pt}} \label{table:vgg6a_ablation-2} 
\end{table}

\paragraph{Diversity of features and Stop-Gradient} In order to improve performance of the overall model, we permit forward communication between within-layer modules. Specifically we allow softmax activations and gradient-decoupled pre-auxiliary module features to be sent across local modules. This corresponds to the stop-gradient and diversity-based approaches discussed in Sec \ref{stop-grad} and \ref{div}. In Table~\ref{table:vgg6a_ablation-1}, we show the ablations which demonstrate that these additions consistently improve the performance of the model allowing it to nearly recover the DGL performance.  

\paragraph{Increasing Width} We also investigate the effect of increasing width in Table~\ref{table:vgg6a_ablation-2}, we observe that increasing the width of a 6-group GN-DGL can bring the performance closer to the DGL performance.

\paragraph{Widthwise splitting of VGG6a} \label{splitting_config_vgg}
The neuron groups within each layer have approximately same number of neurons. For the 2-Group configuration, we split first three convolutional layers in a (2,4,2) pattern respectively, meaning that first convolutional layer is divided in 2 neuron groups, the second into 4, and the third into 2. Furthermore, we introduce G-Group configurations for $G \in \{4,6,8,10\}$, where the layers follow the splitting pattern of (2,4,2) multiplied by $G/2$. i.e., for 6-Group configuration, the split is (6, 12, 6).

\subsection{Multi-layer Grouped Neuron DGL}
\label{sec:ml-gn-dgl}
\begin{figure}[t]
    \centering
    \includegraphics[width=0.85\textwidth,trim=2cm 0cm 1.5cm 0cm]{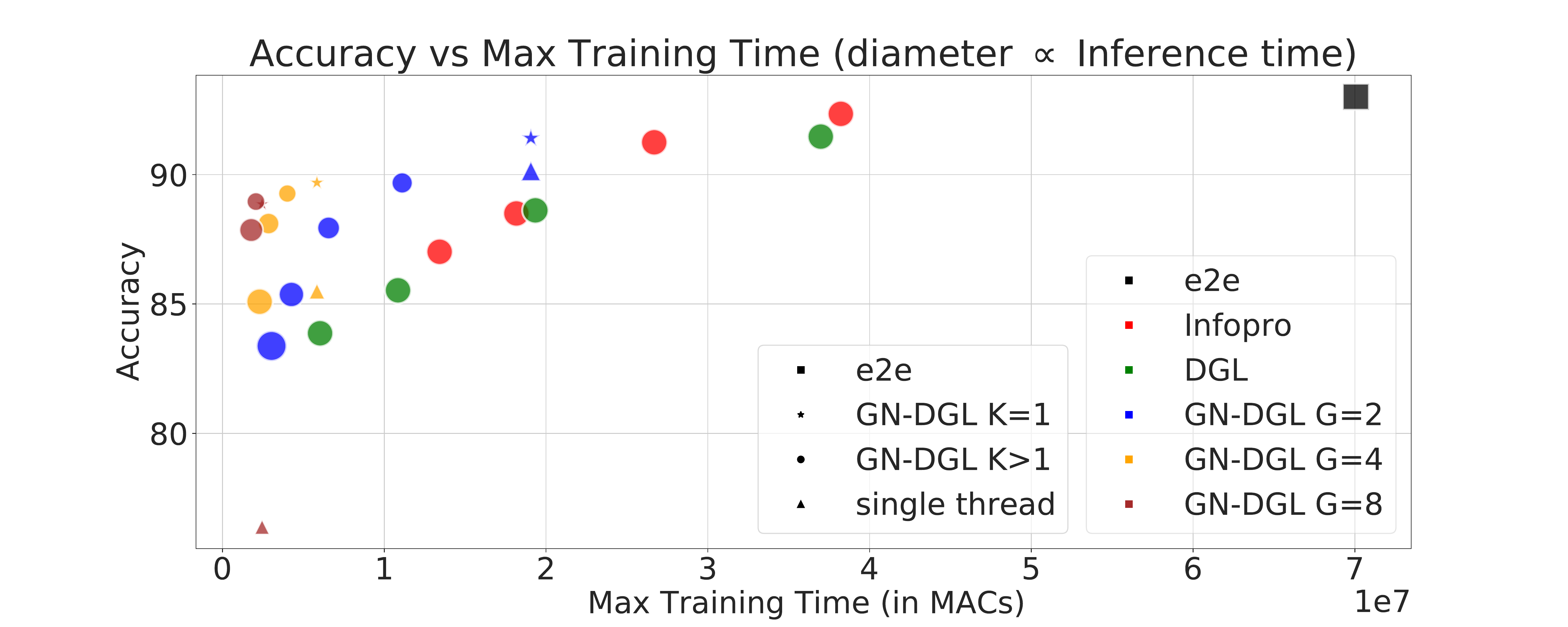}
    \includegraphics[width=0.85\textwidth,trim=2cm 0cm 2cm 0cm]{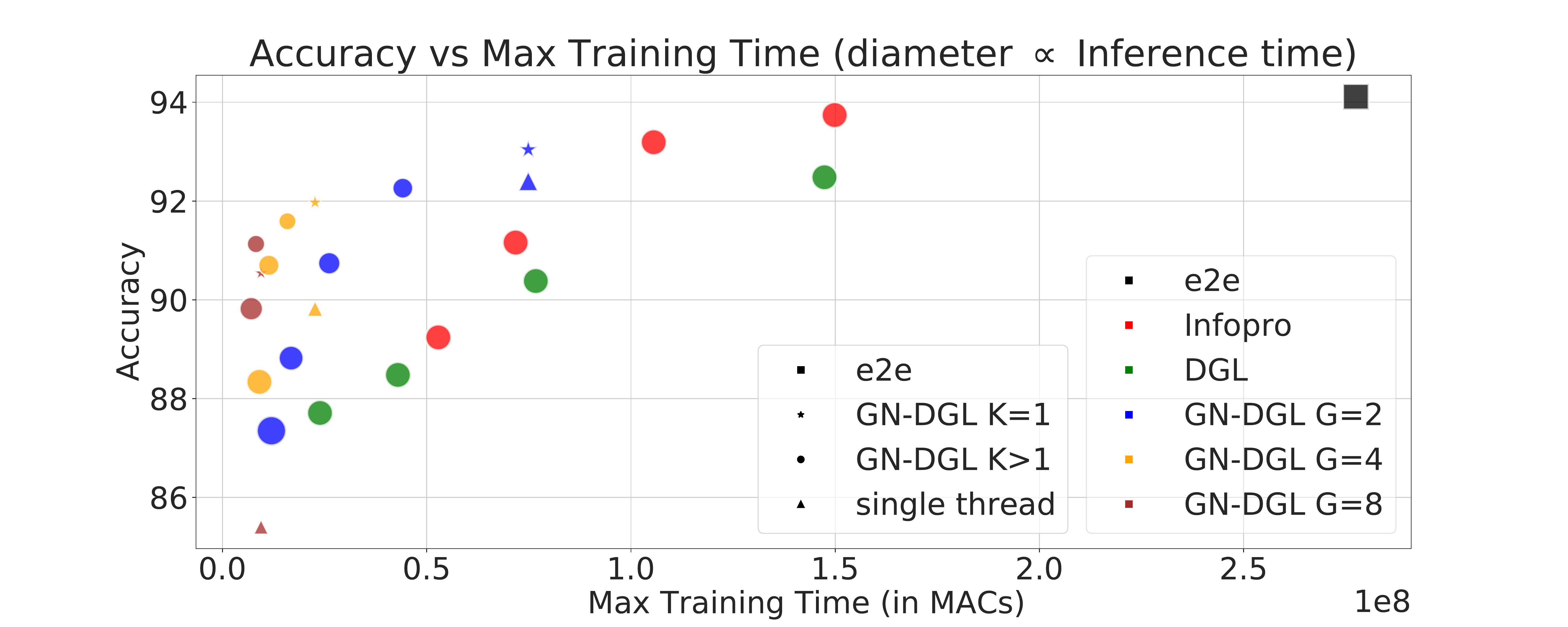}

    \caption{CIFAR-10 results. Accuracy versus maximum training time in MACs for a given node in a theoretical distributed scenario for various local learning methods. We observe that for both ResNet-32 (top) and a wider ResNet-32x2 (bottom), the GN-DGL leads to better tradeoffs in training time versus accuracy. The size of the bubbles is proportional to the inference time of the models.\vspace{10pt}} \label{fig:main_result_resnet_32} 
\end{figure}
\begin{figure}[t]
    \centering
    \includegraphics[width=0.85\textwidth,trim=2cm 0cm 1.5cm 0cm]{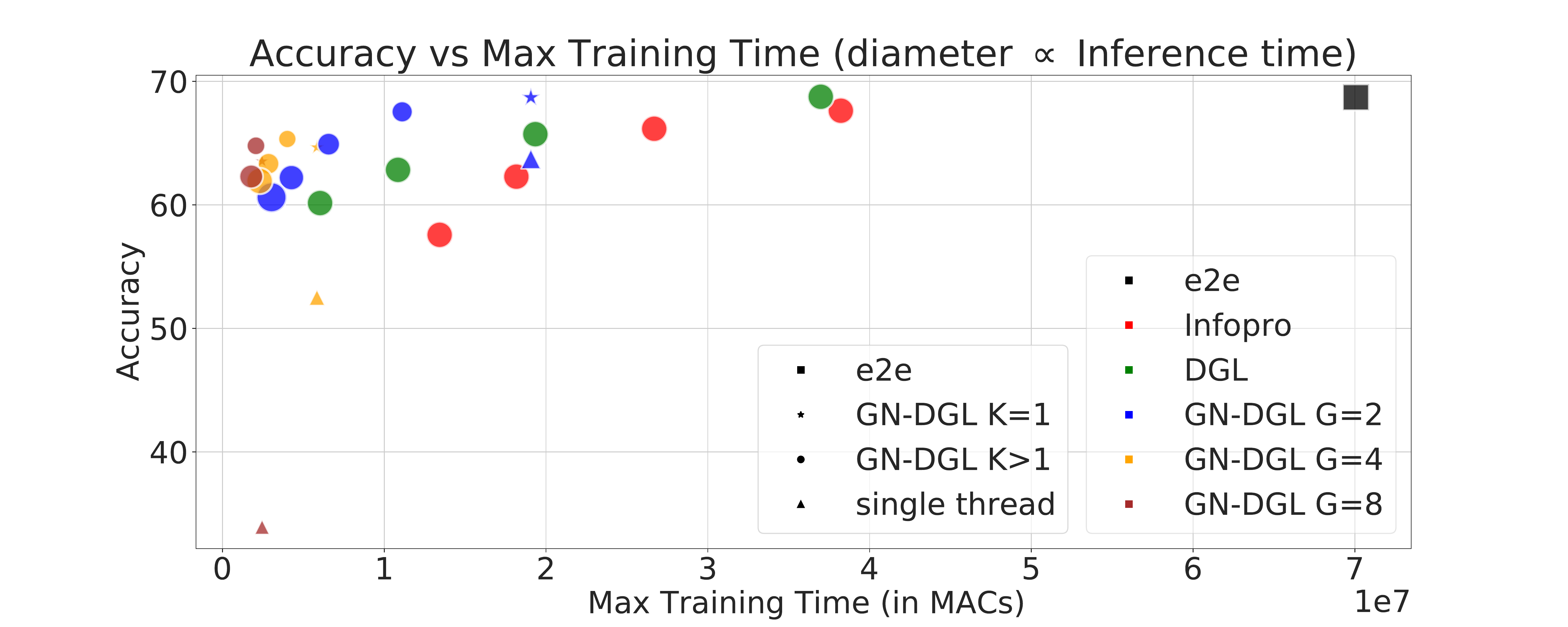}
    \includegraphics[width=0.85\textwidth,trim=2cm 0cm 2cm 0cm]{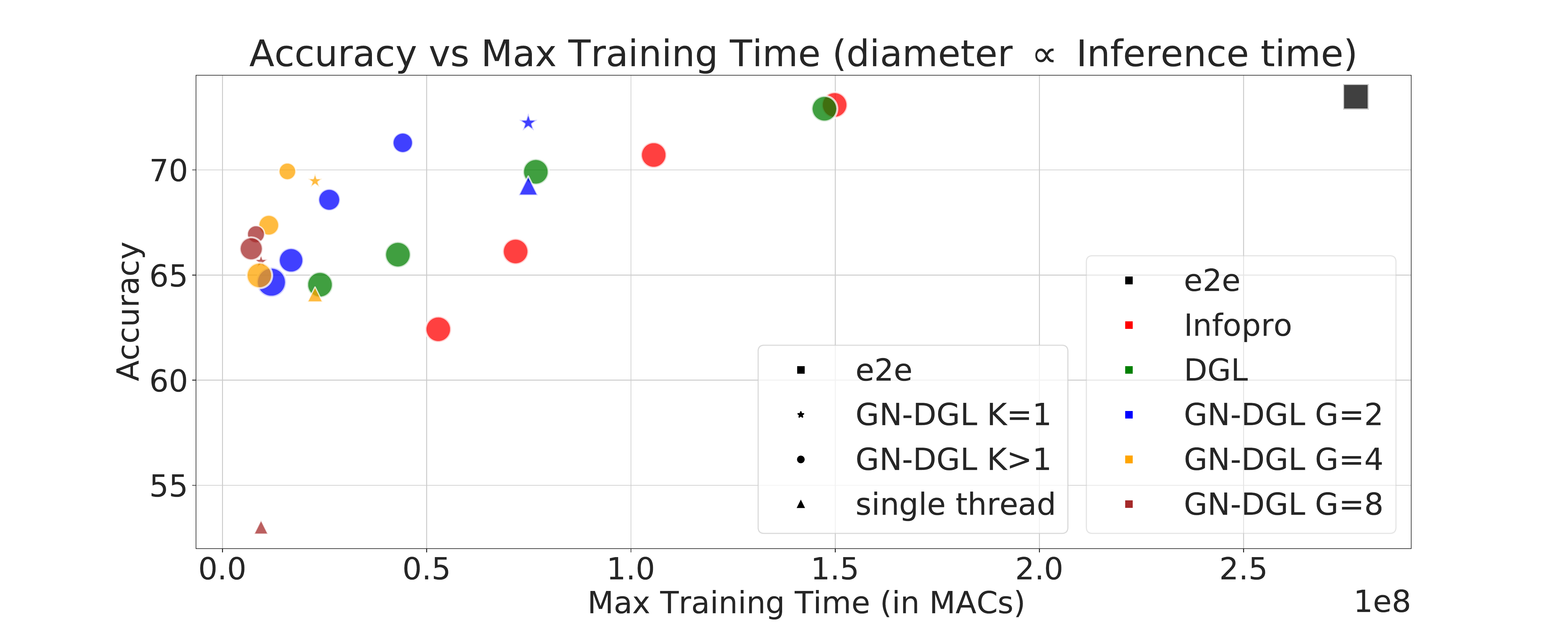}
    \caption{CIFAR-100 results. Accuracy versus maximum training time in MACs for a given node in a  distributed scenario for various local learning methods. We observe that for both ResNet-32 (top) and a wider ResNet-32x2 (bottom), the GN-DGL leads to better tradeoffs in training time versus accuracy, with trends substantially improving over DGL and InfoPro.\vspace{10pt}} \label{fig:main_result_cifar100} 
\end{figure}
\begin{figure}[t]
    \centering
    \includegraphics[width=0.85\textwidth,trim=2cm 0cm 1.5cm 0cm]{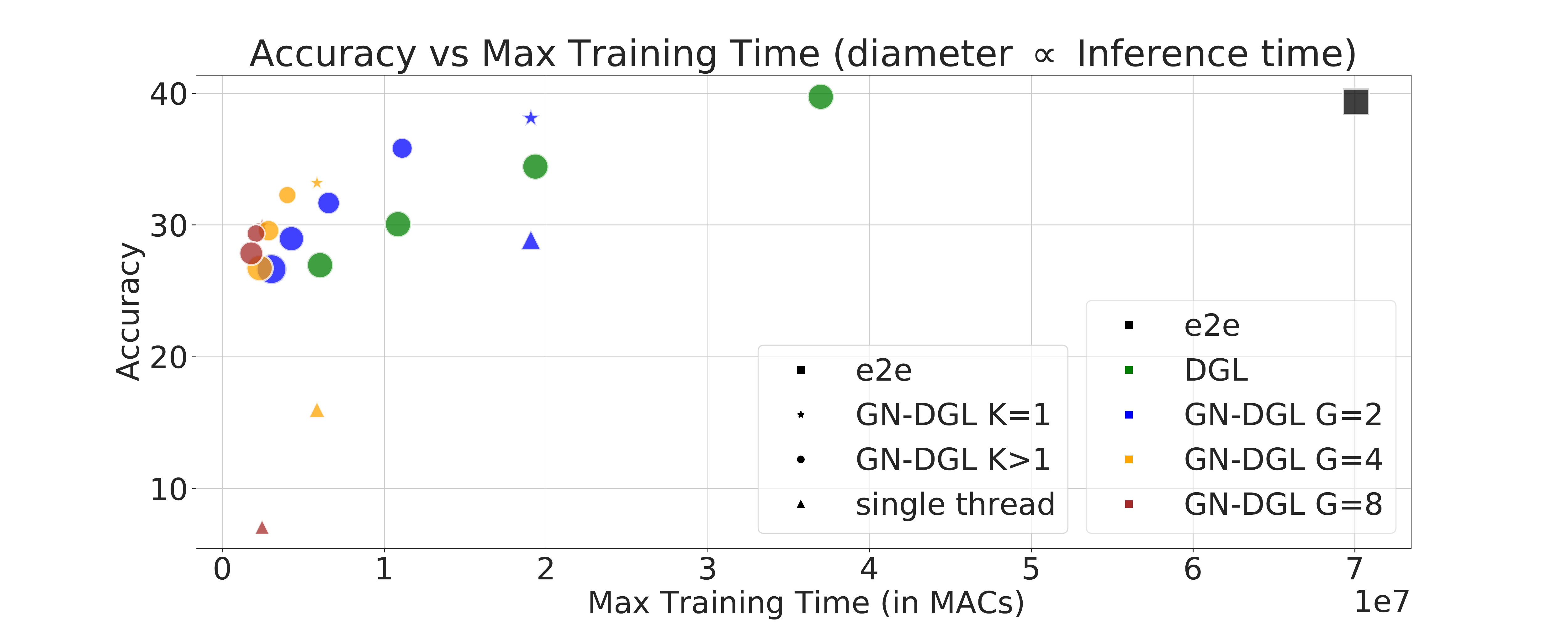}
    \includegraphics[width=0.85\textwidth,trim=2cm 0cm 2cm 0cm]{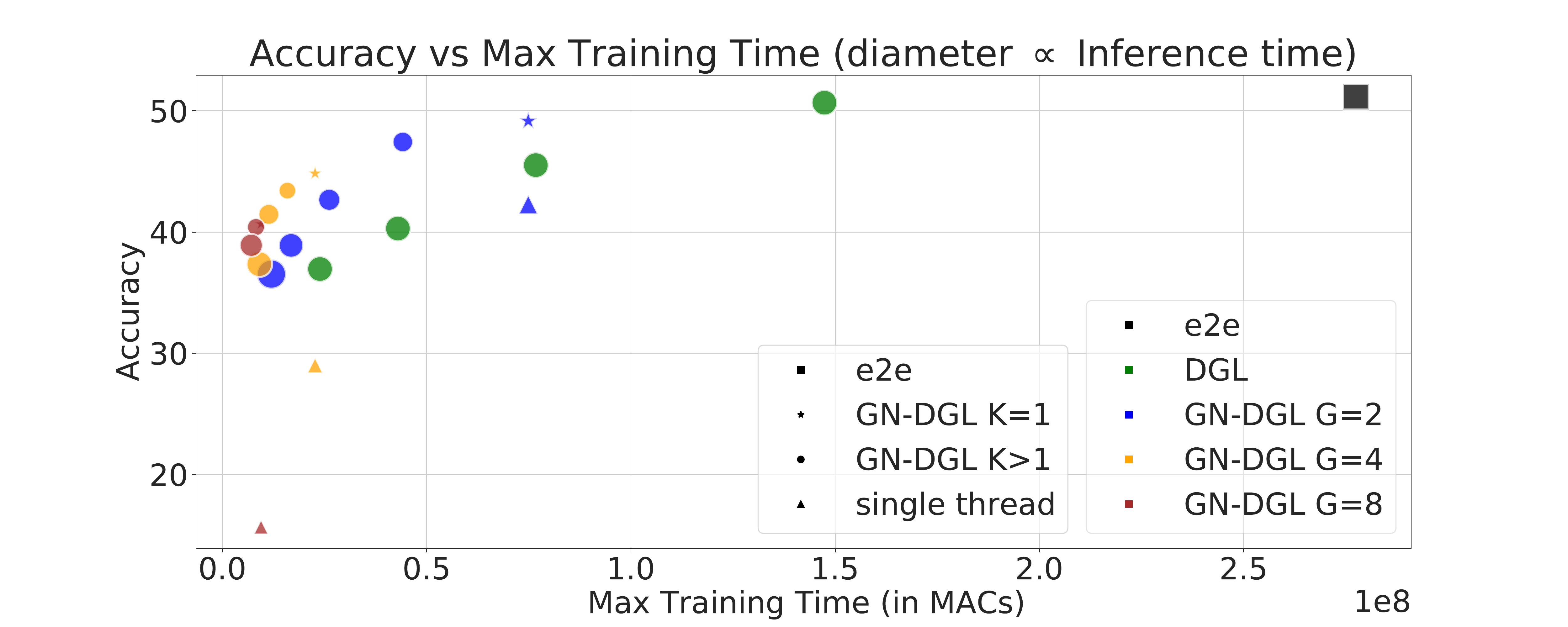}
    \caption{Imagenet32 results. Accuracy versus maximum training time in MACs for a given node in a theoretical distributed scenario for various local learning methods. We observe that for both ResNet-32 (top) and a wider ResNet-32x2 (bottom), the GN-DGL leads to better tradeoffs for this complex dataset. InfoPro is not shown as it yielded very poor performance on this dataset.\vspace{0pt}} \label{fig:main_result_imagenet} 
\end{figure}

Fully layerwise training can lead to performance degradation as found in \cite{pmlr-v119-belilovsky20a,wang2021revisiting}. Thus in this section we extend our splitting techniques to modules, which are defined as collections of layers, and sub-modules: sub-networks that can be grouped together to form a module. Unlike the previous section, for multi-layer versions of GN-DGL, we divide the network in such a way that we get $K$ and $G$ local modules depth-wise and width-wise respectively, resulting in a total of $K \times G$ local modules. Here, $K$ corresponds to depth-wise splitting used in \cite{wang2021revisiting}, on whose experimental setup we base this experiment set. Each local module consists of approximately same number of layers and channels per layer. They are trained in isolation to each other as shown in Figure \ref{fig:ML_GN_DGL}. We note that when performing multi-layer splits in the context of the CNN, the inference time of the overall network is decreased (fewer cross-connections lead to fewer computations). This can provide an additional benefit in applying GN-DGL.

\textbf{Discussion of Baseline Methods}
Here, we compare the proposed approach with other local-learning methods such as Decoupled Greedy Learning (\textit{DGL}) \cite{pmlr-v119-belilovsky20a} and local learning with Information Propagation (\textit{InfoPro}) \cite{wang2021revisiting}. The other local-learning approaches only split the network across depth, unlike our approach of splitting across depth as well as width. Such splitting across both directions gives several advantages in performance and computation speed. We consider other local-learning methods as a special case of G = 1 for comparison. We note an inefficiency of the InfoPro method is its decoder model that tries to reconstruct the input to the network at each layer, leading to high computational cost at deeper layers (due to spatially upsampling a high number of channels). In addition we provide as reference the performance of a model when it is divided in width by $G$ and trained end to end. We denote this baseline as \textit{single thread}.  We also report the performance of the model corresponding to G with K=1, which serves as a baseline of running individual sub-networks trained independently and recombined by a fully connected layer. Finally we compare to end-to-end training ($e2e$).

\textbf{Metrics} In our setup, we consider the three factors for comparison,  total training time, inference time, and final performance as these yield a number of trade-offs. For a fair consideration of the training time, we consider the maximum computation time of any sub-component for a method in multiply–accumulates (MACs).


\paragraph{Discussion of results} Results are presented for CIFAR-10, CIFAR-100, and Imagenet-32 datasets in Figure~\ref{fig:main_result_resnet_32},~\ref{fig:main_result_cifar100},~\ref{fig:main_result_imagenet} respectively. We consider results for ResNet-32 and a wider ResNet-32x2. It can be observed that various configurations of our GN-DGL method have better tradeoffs in many regimes in terms of performance, total training time and inference speed when compared to DGL and InfoPro with ResNet-32. For more complicated datasets CIFAR-100 and Imagenet32, we observe the tradeoffs are more significantly improved with the G$>$1  trending significantly higher than DGL or InfoPro. 

The naive baselines of single thread heavily underperforms in terms of training time and accuracy tradeoffs. We note the simple K=1 approach can perform reasonably for wider models but degrades as G increases, particularly for CIFAR-100 as shown in Figure \ref{fig:main_result_cifar100}. 

InfoPro gives improved performance over DGL on CIFAR-10, however it does not provide substantial improvements for CIFAR-100. Furthermore, we were unable to obtain reasonable performance on Imagenet32 with the InfoPro, either obtaining overly slow performance or low accuracy. We hypothesize that it requires a deep convolutional decoder model, which leads to drastic slowdown. We also observed it was more sensitive to hyperparameters than the DGL and GN-DGL method. 

\begin{table*}[h]
\centering
\begin{tabular}{|c|c|c|c|c|c|c|c|c|c|} 
\hline
\multirow{2}{*}{\textbf{Evaluation Method}} & \multicolumn{4}{|c|}{\textbf{G=2}} & \multicolumn{3}{|c|}{\textbf{G=4}} & \multicolumn{2}{|c|}{\textbf{G=8}}\\ 
\cline{2-10} &

\textbf{K=2} & \textbf{K=4} & \textbf{K=8} & \textbf{K=16} & \textbf{K=2} & \textbf{K=4} & \textbf{K=8} & \textbf{K=2} & \textbf{K=4}\\ 
\hline
last-layer & 89.45 & 87.52 & 84.73 & 82.84 & 89.01 & 88.25 & 84.64 & 88.67 & 87.11\\ 
\hline
layerwise ensemble & 89.68 & 87.94 & 85.37 & 83.38 & 89.27 & 88.11 & 85.09 & 88.96 & 87.86\\ 
\hline

\end{tabular}
\caption{Ablating the effect of layerwise ensembling in GN-DGL on CIFAR-10 with ResNet-32. We observe that layerwise ensembling for GN-DGL yields better performance compared to last-layer evaluation. Performance gains are increased gradually as K increases for a given G.\vspace{5pt}} \label{table:layerwise_ens_ablation-3} 
\end{table*}
\subsection{Ablations of Layerwise Ensembling}

In contrast to traditional methods of using only the last-layer outputs to make the final prediction,  we ensemble the outputs of auxillary modules, as described in \cite{belilovsky2019greedy}. This approach, referred to as layerwise ensembling, is discussed in detail in Section \ref{layerwise-ens-eval}. An ablation study comparing the effectiveness of layerwise ensembling and last-layer evaluation is presented in Table \ref{table:layerwise_ens_ablation-3}. Our results demonstrate that layerwise ensembling consistently leads to improved performance, at no additional cost. For example, we observed a test accuracy of 87.86\% using layerwise ensembling compared to 87.11\% using last-layer evaluation for GN-DGL (K=4, G=8) on CIFAR-10.

 
\section{Conclusions}
We investigated training independent local neuron groups at multiple layers of a neural network. Specifically we studied a common image classification task, focusing on the CNNs. Our results suggest this approach can allow for increased parallelization of local learning methods, unlocking substantial improvements for accuracy given training and inference time. Future investigation may consider more efficient auxiliary models and extensions to additional tasks. 

\section*{Acknowledgements}
This research was funded by internal funding Concordia. We acknowledge resources provided by Compute Canada and Calcul Quebec.
\bibliography{example_paper}
\bibliographystyle{icml2022}

\end{document}